%% file: main.tex
\documentclass[conference]{IEEEtran}
\IEEEoverridecommandlockouts
\usepackage{amsmath,amssymb,amsfonts}
\usepackage{algorithmic}
\usepackage{graphicx}
\usepackage{textcomp}
\usepackage{xcolor}
\def\BibTeX{{\rm B\kern-.05em{\sc i\kern-.025em b}\kern-.08em
    T\kern-.1667em\lower.7ex\hbox{E}\kern-.125emX}}
\usepackage[numbers, sort&compress]{natbib}
\usepackage[edges]{forest}

\tikzset{%
    parent/.style =          {align=center,text width=2cm,rounded corners=3pt, line width=0.3mm, fill=gray!10,draw=gray!80},
    child/.style =           {align=center,text width=2.3cm,rounded corners=3pt, fill=blue!10,draw=blue!80,line width=0.3mm},
    grandchild/.style =      {align=center,text width=2cm,rounded corners=3pt},
    greatgrandchild/.style = {align=center,text width=1.5cm,rounded corners=3pt},
    greatgrandchild2/.style = {align=center,text width=1.5cm,rounded corners=3pt},    
    referenceblock/.style =  {align=center,text width=1.5cm,rounded corners=2pt},
    pretrain/.style =           {align=center,text width=1.8cm,rounded corners=3pt, fill=blue!10,draw=blue!80,line width=0.3mm},   
    pretrain_work/.style =           {align=center, text width=5cm,rounded corners=3pt, fill=blue!10,draw=blue!0,line width=0.3mm},  
    template/.style =           {align=center,text width=1.8cm,rounded corners=3pt, fill=red!10,draw=red!80,line width=0.3mm},   
    template_work/.style =           {align=center,text width=5cm,rounded corners=3pt, fill=red!10,draw=red!0,line width=0.3mm},    
    answer/.style =           {align=center,text width=1.8cm,rounded corners=3pt, fill= cyan!10,draw= cyan!80,line width=0.3mm},   
    answer_work/.style =           {align=center,text width=5cm,rounded corners=3pt, fill= cyan!10,draw= cyan!0,line width=0.3mm},      
}

\begin{document}

\title{A Survey on Prompting Techniques in LLMs}

\author{\IEEEauthorblockN{Prabin Bhandari}
\IEEEauthorblockA{\textit{Department of Computer Science} \\
\textit{George Mason University}\\
Fairfax, Virginia, USA \\
pbhanda2@gmu.edu}
}
\maketitle

\begin{abstract}
Autoregressive Large Language Models have transformed the landscape of Natural Language Processing.
Pre-train and prompt paradigm has replaced the conventional approach of pre-training and fine-tuning for many downstream NLP tasks.
This shift has been possible largely due to LLMs and innovative prompting techniques.
LLMs have shown great promise for a variety of downstream tasks owing to their vast parameters and huge datasets that they are pre-trained on.
However, in order to fully realize their potential, their outputs must be guided towards the desired outcomes.
Prompting, in which a specific input or instruction is provided to guide the LLMs toward the intended output, has become a tool for achieving this goal.
In this paper, we discuss the various prompting techniques that have been applied to fully harness the power of LLMs.
We present a taxonomy of existing literature on prompting techniques and provide a concise survey based on this taxonomy.
Further, we identify some open problems in the realm of prompting in autoregressive LLMs which could serve as a direction for future research.
\end{abstract}


\input{sections/introduction}
\input{sections/preliminaries}

\input{sections/area-taxonomy}
\input{sections/taxonomy-based-survey}

\input{sections/open-problems}

\input{sections/conclusion}

\footnotesize{
\bibliography{bibi}
}
\end{document}

%% file: sections/introduction.tex
\section{Introduction}

Language models (LMs) have long been the de facto standard for modeling natural languages, designed with the purpose of comprehending and generating human-like language.
Language models also serve as the cornerstone upon which a multitude of Natural Language Processing (NLP) applications are built, including but not limited to machine translation, sentiment analysis, document classification, and chatbots.
This has made LMs one of the most extensively researched domains within the field of NLP.
Initially, language modeling relied on statistical approaches, which eventually gave way to neural language models fueled by the increase in our computing resources.
With neural LMs came the requirement for substantial labeled data, as any downstream task was accomplished via supervised training.
The introduction of Transformers~\cite{vaswani2017attention} marked a pivotal shift, introducing the pre-train and fine-tune paradigm of NLP.
Transformers gave rise to pre-trained language models (PLMs).
These PLMs, while still being neural LMs, mostly used architecture similar to the transformer or a variant of it.
The key difference between neural LMs and PLMs is in their training process.
PLMs are pre-trained on a vast amount of textual data in a semi-supervised approach, eliminating the need for labeled data during pre-training.
Following their pre-training, PLMs are fine-tuned for specific downstream tasks using relatively less labeled data than what neural LMs typically require.
This is because PLMs have a strong foundation in language modeling through pre-training.
Recently, the pre-train and fine-tune paradigm has evolved into a pre-train and prompt approach, mainly due to the emergence of Large Language Models~(LLMs).

LLMs evolved from PLMs as researchers tried to enhance the performance of PLMs by increasing the model's size, dataset volume, and/or computational resources employed during training.
This drive was motivated by scaling law~\cite{kaplan2020scaling}.
In doing so, the performance of these models did indeed increase as expected.
However, it also led to the rise of emergent abilities~\cite{wei2022emergent} that cannot be explained solely by scaling law.
Emergent abilities refer to the capabilities of an LLM that manifest when it reaches a certain scale, achieved through an increase in the number of parameters, pre-training data, computational resources, or a combination of these factors.
The majority of LLMs nowadays are autoregressive in nature.
This means that they generate the next token based on the previous tokens only.
These autoregressive LLMs use either the encoder-decoder architecture or only the decoder of the transformer.
Given that these LLMs have been pre-trained on a plethora of text and often the whole internet, they are an extremely powerful tool.
However, harnessing their immense potential could be challenging due to their large size, making it infeasible to fine-tune them for each downstream task.
This challenge has been mitigated by the emergence of prompting techniques in LLMs, which seek to replace the need for fine-tuning LLMs for each downstream task individually.

Prompting refers to providing specific instruction or input, whether in human-readable form or not, to the LM in order to accomplish the downstream task.
The use of prompting first appeared in the context of PLMs.
For instance, \citet{petroni-etal-2019-language} used prompting when attempting to quantify the amount of world knowledge embedded in PLMs.
Their hypothesis was that since PLMs were pre-trained on vast amounts of data, PLMs might contain significant world knowledge embedded within their parameters.
Concretely, their input to the PLM was: \textit{ Francesco Bartolomeo Conti was born in [MASK].}.
Since they employed a masked language model in their experiments, the expectation was that the model would replace the \textit{[MASK]} token with the correct answer.
Prompting with respect to autoregressive LLMs was popularized by~\citet{brown2020language}.
They demonstrated an emergent ability of LLMs, facilitated by prompting techniques, known as in-context learning.
Basically, their research shows that LLMs, when provided with a task instruction and a few demonstration examples, could handle a wide range of downstream NLP tasks.
This novel prompting technique for LLMs eliminated the need to fine-tune an LLM and opened a new field of research into prompting techniques for LLMs.

Given the recent proliferation in prompting techniques, this paper discusses the current literature concerning prompting techniques tailored for autoregressive LLMs.
The structure of the paper is as follows.
Section II lays the groundwork by discussing the preliminaries required to follow the rest of the paper.
Section III introduces the taxonomy used to categorize prompting techniques in autoregressive LLMs.
Section IV provides a concise survey of existing literature, organized according to the established taxonomy.
Section V discusses the prevailing challenges and open problems in the field of prompting, offering potential future research directions.
Finally, we conclude with our remarks in Section VI.

%% file: sections/preliminaries.tex
\section{Preliminaries}

\subsection{Language Models}

A language model~(LM) is a computational model of a natural language that predicts and generates human language based on a text corpora that it was trained on.
Technically, LMs model the likelihood of a sequence of tokens in order to predict the probabilities for the missing or next token.
Language models can broadly be classified into four main development stages: 

\subsubsection{Statistical language models (SLM)}
SLMs, introduced in the early 1980s, are based on statistical learning methods.
They assign probabilities to tokens based on a statistical model of a text corpora.
One of the notable SLMs is called the N-gram model.
The N-gram models are based on Markov's assumption that the probabilities of the next token are based on the last \((N-1)\) tokens.
Formally,
\begin{equation}
\begin{split}
  P(w_n | w_{n-1:1}) = P (w_{n-1}|w_{n-2:1} * P (w_{n-2}|w_{n-3:1}) \\
    * ... *P (w_2|w_1 ) * p(w_1)  
\end{split}
\end{equation}

The probabilities are estimated using a Maximum Likelihood Estimate (MLE) based on the frequency of the tokens in the text corpus.
Along with n-gram models, the Hidden Markov Model (HMM) based language models, and maximum entropy-based language models were some of the other popular SLMs. 

\subsubsection{Neural language models (NLM)}
NLMs leverage neural networks to estimate token probabilities.
While various neural network architectures, such as feed-forward neural networks (FFNN) and convolutional neural networks (CNN), have been explored for language modeling, recurrent neural networks (RNN) emerged as the dominant choice.
However, the landscape underwent a significant transformation with the introduction of Transformers~\cite{vaswani2017attention}.

Transformer, originally purposed for machine translation tasks, features an encoder-decoder architecture.
In this design, the encoder encodes the input token sequence to extract context, while the decoder's main role is to generate the next token based on the encoder's encoded information and already decoded tokens.
Transformers introduced a technique called self-attention which captures complex contextual relationships within sequences.
The original transformer used multi-headed self-attention, enabling the model to attend to different parts of the input sequence simultaneously, significantly boosting its capacity to model complex dependencies.

\subsubsection{Pre-trained language models (PLM)}

Earlier efforts for PLMs involved training shallow networks for word embeddings, like word2vec~\cite{mikolov2013efficient} and Glove~\cite{pennington2014glove}, to capture the semantic meaning of the word.
Although these models were highly effective, they failed to capture essential contextual information related to a word.
Subsequent models like ELMo~\cite{sarzynska2021detecting}, employed RNNs in an effort to gather contextual information in the word embeddings; however, they were constrained by their size.

With the introduction of transformers, researchers were able to train more deeper networks.
Further, the advent of self-supervised learning techniques~\cite{liu2021self} meant that costly human-annotated data are not required to train the LMs.
In the self-supervised learning paradigm, the LMs are pre-trained with the help of raw textual data only, completely bypassing the need for human-annotated data.
These PLMs are often then fine-tuned for specific downstream tasks with human-annotated data.
However, due to the effectiveness of pre-training, the requirement for the size of human-annotated data drops off.

Various pre-training objectives are used with alterations to the standard encoder-decoder architecture which gives rise to three different types of PLMs: (a) Left-to-right PLM, (b) Masked PLM, and (c) Encoder-decoder PLM.
Figure~\ref{fig:plm_models}, extracted from~\cite{wang2022pre}, illustrates the architectures of these three different PLMs.
Further explanation of these PLMs is provided below:

\begin{figure*}[htbp]
\centerline{\includegraphics[width=\linewidth]{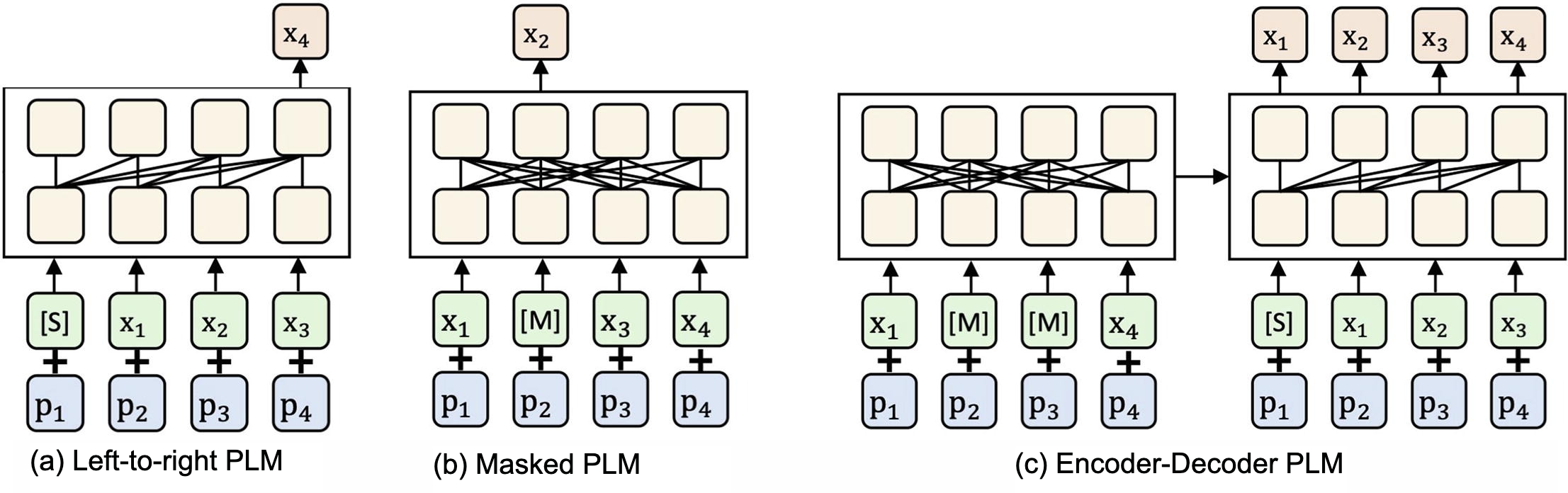}}
\caption{Architecture of different PLMs. Image from~\cite{wang2022pre}.}
\label{fig:plm_models}
\end{figure*}

\paragraph{Left-to-Right PLM}
Left-to-right (or similarly Right-to-left) PLMs, like GPT~\cite{radford2018improving} and its successor GPT-2~\cite{radford2019language}, are autoregressive language models that usually employ the decoder part of the original transformer architecture and are pre-trained to predict the upcoming tokens one at a time based on the previous tokens using a large text corpus.

\paragraph{Masked PLM}
Masked PLMs are non-autoregressive in nature meaning that they consider both the preceding and following tokens when predicting token probabilities. 
Masked PLMs generally use the encoder part of the original encoder-decoder architecture of the transformer.
These PLMs are pre-trained by applying a noising function that corrupts the text and the pre-training objective is to restore the original uncorrupted text.
An early example of such a PLM is BERT~\cite{devlin-etal-2019-bert}, where certain tokens within a sentence are replaced with \textit{[MASK]} and the pre-training objective was to predict the \textit{[MASK]} token.
Other various noising functions have also been proposed~\cite{JMLR:v21:20-074, liu-etal-2020-multilingual-denoising, lan2019albert}.

\paragraph{Encoder-Decoder PLM}
Encoder-Decoder PLMs use the original architecture of the transformer.
While they are generally auto-regressive in nature, there are a few models purposed with non-autoregressive design.
Typically, a noising function is applied to the encoder, while a standard language modeling is used at the decoder part during the model's pre-training.
Notable examples of Encoder-Decoder PLMs include Meta's BART~\cite{lewis-etal-2020-bart} and Google's T5~\cite{2020t5}.

\subsubsection{Large language models (LLM)}

The scaling law~\cite{kaplan2020scaling} dictates that an increase in the model's size, dataset size, and computational resources used during training often results in enhanced performance on the downstream tasks.
Researchers have tried constantly to push the boundaries of this law by continually increasing the model's size.
For instance, GPT-3~\cite{brown2020language} has a massive 175 billion parameters, while PaLM~\cite{chowdhery2022palm} surpasses even that with 540 billion parameters.
Despite having similar training methodologies compared to other PLMs, these large PLMs exhibit \textit{emergent abilities}~\cite{wei2022emergent}.
For example, GPT-3 can learn a task description with the help of a few examples passed as context whereas the predecessor of GPT-3, GPT-2 can not do that.
In contemporary times, the term ``Large language models (LLMs)" primarily refers to these massive language models, having tens or even hundreds of billions of parameters, and trained on vast datasets.
These LLMs predominantly adopt Left-to-right transformer architecture (decoder-only), and they commonly exhibit an autoregressive nature.

\subsection{Prompting}
Prompting refers to providing a specific input or instruction to guide the model's output.
Basically, an input \(x\) with the help of a prompt template \(f\) is converted to a new representation \(f(x)\), which is then fed into the model to get the desired output \(y\).
We generally employ two kinds of prompts, cloze prompts, and prefix prompts.

Cloze prompts are popular with masked language models, where the objective is to fill in the blanks.
For example, if our task is to find the capital city of a country, our cloze-style prompt will be as follows:
\begin{verbatim}
The capital of Nepal is [BLANK].
\end{verbatim}

The model is expected to fill the blank with the correct answer.

Prefix prompts are generally employed with autoregressive LLMs, where the goal is for the model to produce the continuation of the string input.
For example, if our task is to convert a sentence in English to a sentence in Nepali, our prefix-style prompt will be as follows:
\begin{verbatim}
Convert the following English sentences to
Nepali:
English: All the world’s a stage, and all 
the men and women merely players.
Nepali:

\end{verbatim}
The model is expected to produce a continuation of this input where the output will be the Nepali translation of the input English sentence.

If we provide these templates without additional examples, it is referred to as zero-shot prompting.
However, if we provide a few illustrative examples of the correct inputs and outputs, it is referred to as few-shot prompting.

%% file: sections/area-taxonomy.tex
\section{Area Taxonomy}

\input{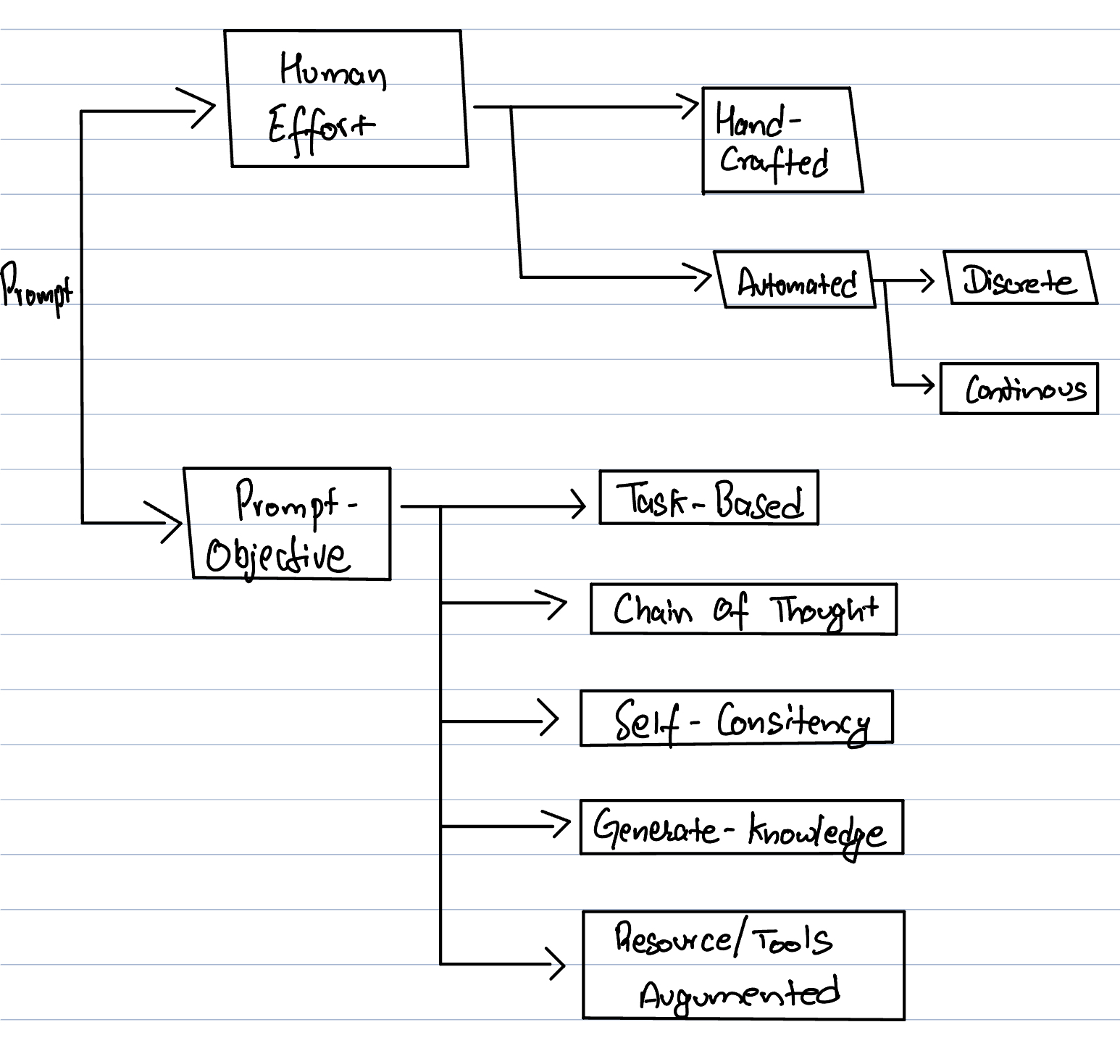}
Figure~\ref{fig:taxonomy} presents the area taxonomy of prompting methods in autoregressive LLMs.
The classification of prompts is based on two key dimensions: the level of human involvement in prompt creation and the specific types of these prompts.
In terms of human effort, prompts are categorized into two groups: ``Hand-Crated" and ``Automated", reflecting the extent of manual input required in the prompt creation process.
Additionally, prompts are categorized into three distinct groups based on their intended objectives.
These categories include ``Task-Based", `Generate-Auxiliary" and, ``Resource/Tools Augmented".
It is important to note that this classification is based on the goals and purpose of the prompts themselves, rather than the ultimate objective of the downstream tasks.
In the next section, we delve into existing research within each of these classifications to provide a comprehensive overview of the field.

%% file: sections/taxonomy.tex
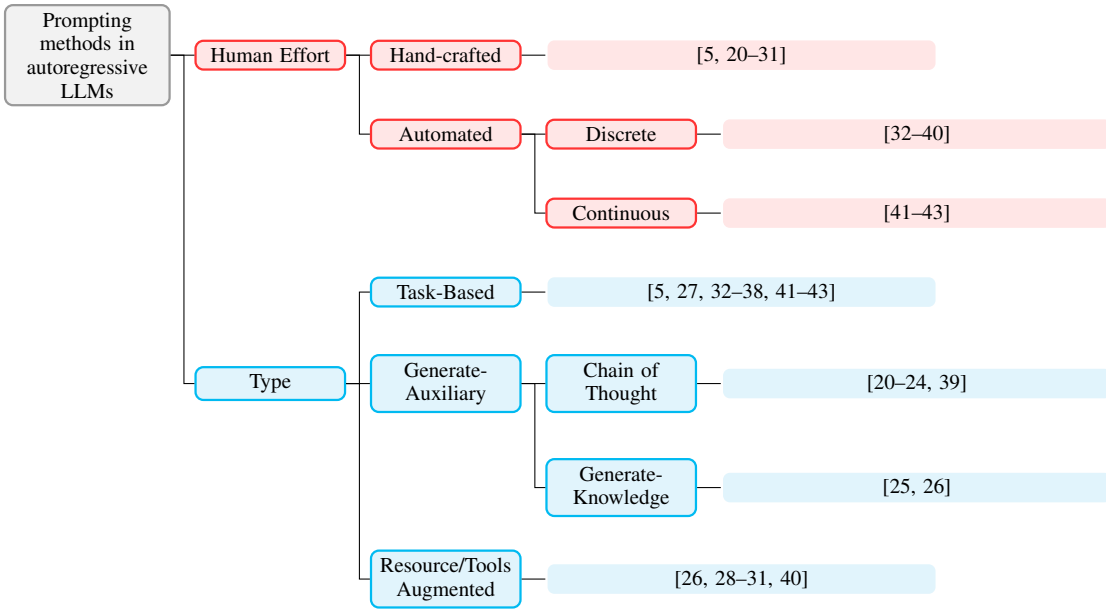
\begin{figure*}
\footnotesize
        \begin{forest}
            for tree={
                forked edges,
                grow'=0,
                draw,
                rounded corners,
                text width=4.7cm,
                s sep=17pt,
                calign=child edge, 
                calign child=(n_children()+1)/2,
            }
            [Prompting methods in autoregressive LLMs, parent
                [Human Effort, template
                    [Hand-crafted, template
                        [\cite{brown2020language, wei2022chain, kojima2022large, wang2022self, yao2023tree, tree-of-thought-prompting, liu2021generated, press2022measuring, zhou2022least, gao2023pal, chen2022program, long2023large, yao2022react}, template_work]                            
                    ]
                    [Automated, template
                        [Discrete, template
                            [\cite{jiang2020can, yuan2021bartscore, haviv-etal-2021-bertese, wallace-etal-2019-universal, davison-etal-2019-commonsense, lu-etal-2022-fantastically, deng-etal-2022-rlprompt, zhang2022automatic, paranjape2023art}, template_work]                            
                        ]
                        [Continuous, template
                            [\cite{li-liang-2021-prefix, lester-etal-2021-power, liu2023gpt}, template_work]                            
                        ]
                    ]
                ]
                [Type, answer
                    [Task-Based, answer
                        [\cite{jiang2020can, yuan2021bartscore, haviv-etal-2021-bertese, wallace-etal-2019-universal, davison-etal-2019-commonsense, lu-etal-2022-fantastically, deng-etal-2022-rlprompt, li-liang-2021-prefix, lester-etal-2021-power, liu2023gpt, brown2020language, zhou2022least}, answer_work]                            
                    ]
                    [
                    Generate-Auxiliary, answer
                        [Chain of Thought, answer
                            [\cite{wei2022chain, kojima2022large, zhang2022automatic, wang2022self, yao2023tree, tree-of-thought-prompting}, answer_work]                            
                        ]
                        [Generate-Knowledge, answer
                            [\cite{liu2021generated, press2022measuring}, answer_work]                            
                        ]
                    ]
                    [Resource/Tools Augmented, answer
                        [\cite{gao2023pal, chen2022program, press2022measuring, long2023large, yao2022react, paranjape2023art}, answer_work]                       
                    ]
                ]
            ]
        \end{forest}
\caption{Area Taxonomy of prompting methods in autoregressive LLMs.}
\label{fig:taxonomy}
\end{figure*}

%% file: sections/taxonomy-based-survey.tex
\section{Taxonomy-Based Survey}

In this section, we offer a survey of the existing literature about prompting within the domain of autoregressive LLMs, structured according to the taxonomy introduced in section 3.
It is noteworthy that while some of the research discussed may not be exclusive to autoregressive LLMs, or may have originally targeted PLMs, many of these approaches are applicable and adaptable for effective use with autoregressive LLMs.
\subsection{Human Effort}

On the basis of the amount of human effort required to create the prompts, they can be classified into Hand-crafted and Automated.

\subsubsection{Hand-crafted Prompts}
Hand-crafted prompts are the most natural way of creating prompts where a prompt template is created based on human intuition.
\citet{brown2020language} introduced hand-crafted prefix prompts for solving a variety of tasks.
We provide an example of a hand-crafted prompt taken from \cite{brown2020language}:
\begin{verbatim}
Translate English to French:
Cheese =>
\end{verbatim}

The prompt above is called a zero-shot prompt as we have only provided the task description along with the input.
If we provide a few input-output examples along with the task description, we call them few-shot prompts.
An example of a few-shot prompt is provided below:

\begin{verbatim}
Translate English to French:
Sea otter => loutre de mer
plush girafe => girafe peluche
Cheese =>
\end{verbatim}

\subsubsection{Automated Prompts}
Although manual prompt creation is intuitive, such hand-created prompts may have limitations.
One such limitation is that these hand-crafted prompts may be sub-optimal.
Another issue is that hand-crafted prompts are domain-specific and it can be an arduous task to hand-craft a prompt for some complex downstream tasks.
Consequently, researchers are exploring automated methods for prompt template design. 
These automatically generated prompts can be further classified into discrete and continuous prompts.
\paragraph{Discrete Prompts}
Discrete Prompts, also referred to as ``hard prompts", are those prompts where the prompt input to the underlying LLM is still an actual text.
These prompts are named discrete prompts because our search space for the prompt is limited to the discrete space of the tokens of the underlying LLM.
Different techniques including mining, paraphrasing, and searching have been explored to generate discrete prompts.

In their work, \citet{jiang2020can} proposed a mining-based approach to find discrete prompts.
Originally proposed for masked language models, this approach is adaptable to autoregressive LLMs.
Given an input-output pair of \(x,y\), the method scraps a large text corpus, identifying strings containing both \(x\) and \(y\), and subsequently extracts the middle word or dependency path between them to determine a relation (\(r\)) for use as a prompt: ``\textit{[x] r ...}".
Additionally, \citet{jiang2020can} also proposed the use of paraphrasing for creating discrete prompts.
The proposed solution was to translate a seed prompt into another language which is back-translated to the original language.
Other paraphrasing-based solutions include synonym substitution using a Thesaurus~\cite{yuan2021bartscore} and employing a neural model for rewriting the prompt~\cite{haviv-etal-2021-bertese}.

\citet{wallace-etal-2019-universal} proposed a gradient-based search technique to find discrete prompts.
They employ a gradient-guided search across tokens of the underlying LLM to identify short trigger sequences capable of inducing the desired output from the LLM.
Some approaches score the prompt using another LM.
For example, \citet{davison-etal-2019-commonsense}, handcrafted a set of potential templates which were filled with input and output data from the training set.
These filled prompt templates were scored using GPT-2, with the highest-scoring prompt being selected for use.
\citet{lu-etal-2022-fantastically} also proposed a scoring-based method to address the problem of order sensitivity within the few-shot setting.
Their approach involves considering all possible ordering permutations for the provided few-shot examples.
They then use the underlying LLM to generate from these permutations to generate a probing set.
The probing set is scored using entropy-based measures to rank the effectiveness of different permutations.

As adaptation of reinforcement learning (RL) continues to grow within LLMs, efforts have been made to leverage RL for the optimization of discrete prompts.
One notable contribution is RLPrompt~\cite{deng-etal-2022-rlprompt} which introduces a parameter-efficient policy network.
This network is trained with rewards to generate optimized discrete prompts.

\paragraph{Continuous Prompts}

Continuous prompts, also referred to as ``soft prompts", are those prompts that are defined in the embedding space of the LLM and therefore are not in human-readable format.
The templates of soft prompts have their own parameters that can be tuned.
These prompts are called continuous because our prompt to the LLM are continuous vectors instead of discrete tokens.
We discuss some of the seminal works below.

Prefix Tuning~\cite{li-liang-2021-prefix} involves the addition of task-specific prefixes to the beginning of input sequences. 
These prefixes are free-parameters (\(P_{\theta}\)), which undergo reparameterized through a small multi-layer perception during the training process.
Then, the log-likelihood objective is optimized,  with the LLM parameters frozen while only updating the prefix parameters.
Mathematically,
\[   \max_{\theta} \log P(y|x; \theta; \phi) = \max_{\theta} \sum_{y_i} \log P(y_i | h_{< i}; \theta; \phi)  \]
where, \(\phi\) are the parameters of the LLM, and \(h_i\) is the contamination of all neural network layers at time step \(i\).
If \(i \in P_{idx}\), we use the prefix parameters otherwise we use the LLM parameters.
Notably, \citet{li-liang-2021-prefix} observed that prefix tuning significantly enhanced effectiveness compared to discrete prompts, especially in the low-data and out-of-domain settings.
A similar approach is prompt tuning~\cite{lester-etal-2021-power}, where special tokens are prepended to the input sequence and these tokens are tuned directly.

P-tuning~\cite{liu2023gpt} uses a hand-crafted prompt where all the tokens except for the input(\(x\)) are considered pseudo-tokens and are mapped to trainable embedding vectors.
They model these trainable embedding vectors through a bi-directional long-short-term memory (LSTM) model.
One added benefit of P-tuning to prefix and prompt tuning is that the continuous vectors can be inserted anywhere as opposed to the beginning of input only.
P-tuning was originally proposed to solve natural language understanding tasks through GPT-based models.

\subsection{Type}

The primary objective of any downstream task is to achieve a specific goal, such as classifying input data or generating contextually appropriate responses.
In classification tasks, the goal is to discriminate between different class labels and assign the proper class label for an input.
The broad objective of aligning with the goals of the downstream task they support is shared by all prompting approaches.
However, in practice, prompts might often serve additional auxiliary purposes or use additional tools to facilitate the downstream task objective.
Based on these additional or auxiliary purposes or tools of prompts, we can classify them into different categories.
It is important to note that these categories may encompass both hand-crafted and automated prompts.
We provide a description of each category below:

\subsubsection{Task-Based}

Task-based prompts are the most straightforward category within the taxonomy of prompts based on their objective.
These prompts do not serve any auxiliary goal and are characterized by their single objective of the downstream task.
All the different prompting techniques that we discussed under Hand-Crafted and Automated prompts fall under this category.

\subsubsection{Generate-Auxiliary}

Generate-Auxiliary prompts are the types of prompts that generate auxiliary output text in order to facilitate the downstream tasks.
Generate-Auxiliary prompts can be further classified into chain of thought and generate-knowledge prompts.

\paragraph{Chain of Thought (CoT)}
\begin{figure}[htbp]
\centerline{\includegraphics[width=\linewidth]{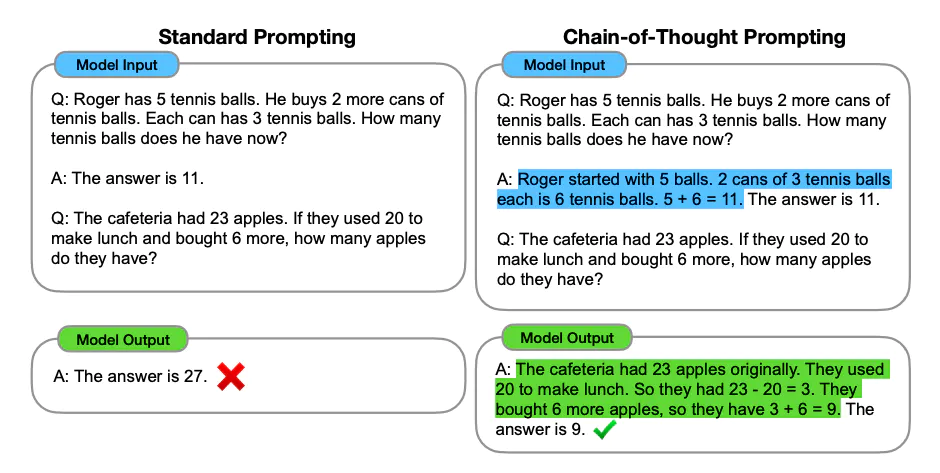}}
\caption{Chain of thought prompting.  Image from~\cite{wei2022chain}.}
\label{fig:cot}
\end{figure}
The use of prompts that elicit a coherent series of intermediate reasoning steps, ultimately leading to the formulation of the final answer, is known as the chain of thought (CoT) prompting.
\citet{wei2022chain} popularized the term ``Chain of thought prompting" in their seminal paper.
Chain of thought prompting finds its usefulness in arithmetic reasoning, commonsense reasoning, and symbolic reasoning tasks.
We present an example of the chain of through prompting from \cite{wei2022chain} in figure \ref{fig:cot}.
As can be seen from the figure, in standard prompting the correct answer may not be readily obtained.
However, when employing the chain of thought prompting, where the answers of few-shot examples also contain the intermediate steps to reach the answer, the model generates similar intermediate reasoning steps and reaches the final correct answer.
Chain of thought is an emergent ability of sufficiently large language models that allows LLMs to perform reasoning tasks. 

Zero-shot CoT~\cite{kojima2022large} can be considered the zero-shot version of CoT purposed by \citet{wei2022chain}.
In zero-shot Cot, ``Let's think step by step" is added to the prompt, and with sufficiently large language models, we get a series of reasoning steps leading to correct answers.
Auto CoT~\cite{zhang2022automatic} alleviates the problem of sub-optimal demonstrations in CoT. 
Auto CoT uses zero-shot CoT to create demonstrations from the LLM itself and uses them in few-shot prompting scenarios similar to CoT.
The key technique behind Auto CoT involves the partition of questions within a given dataset into a few clusters.
From each cluster, a representative question is selected, promoting diversity in task demonstration and ultimately enhancing model performance.

\begin{figure}[htbp]
\centerline{\includegraphics[width=\linewidth]{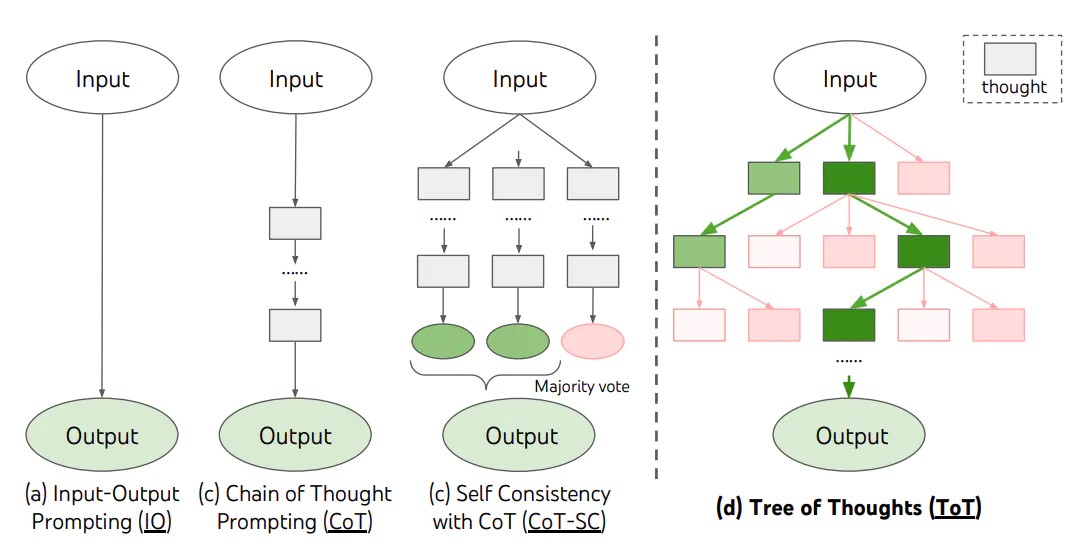}}
\caption{Self-consistency methodology for CoT prompting in comparison to CoT prompting with greedy decoding. Image from~\cite{yao2023tree}.}
\label{fig:tot}
\end{figure}

Self-consistency~\cite{wang2022self} represents a recent and noteworthy idea within the CoT domain.
It is important to note that self-consistency is not inherently a prompting technique but rather a decoding strategy employed when utilizing CoT prompts.
CoT prompts are generally associated with a greedy decoding strategy, mainly because reasoning tasks often have a single fixed answer.
However, the authors argue that introducing diversity into the reasoning process can be highly advantageous.
The self-consistency methodology involves prompting the LLM with CoT prompts, followed by sampling from the LLM to generate a diverse set of reasoning paths, thus deviating from the conventional greedy search approach.
Finally, the method involves marginalizing these reasoning paths to identify and select the most consistent answer as the final output.
Tree-of-Thoughts (ToT)~\cite{yao2023tree} presents an idea similar to self-consistency but with enhancements aimed at refining the self-consistency approach.
In contrast to self-consistency, where a majority voting mechanism determines the final answer, ToT adopts a more explorable tree structure for the thoughts.
ToT maintains a tree comprising individual thought steps, with each thought being self-evaluated by the LLM.
This tree-based framework is complemented by the application of search algorithms, including Breadth First Search and Depth First Search, which facilitates for systematic exploration of thoughts.
Figure \ref{fig:tot}, taken from \cite{yao2023tree}, illustrates various CoT prompting strategies described above.

\citet{tree-of-thought-prompting} proposed a text-based prompt that works similarly to ToT.
The prompt used in \cite{tree-of-thought-prompting} is:
\begin{verbatim}
Imagine three different experts are 
answering this question.
All experts will write down 1 step of
their thinking, then share it with the 
group.
Then all experts will go on to the next 
step, etc.
If any expert realises they're wrong at
any point then they leave.
The question is...
\end{verbatim}

\citet{zhou2022least} argues that the original CoT cannot generalize well to hard problems when the demonstrations provided are easy and to overcome this problem propose Least-to-most prompting.
Least-to-most prompting works in two stages.
Firstly, the given question is decomposed into a series of subquestions by prompting the LLM as: \textit{To solve \(\langle\)Q\(\rangle\), we need to first solve...}.
Then, each subquestion is solved by the LLM to reach the final answer.

\paragraph{Generate knowledge}

\citet{liu2021generated} introduced the concept of `Generated Knowledge Prompting'. 
The basic idea behind this approach is to generate task-specific knowledge, either by leveraging the same LLM used for the downstream tasks or by leveraging a separate  LLM, and subsequently incorporating this knowledge into the prediction process.
Generated knowledge prompting is a two-step process.
First, in a few-shot setting, question-related knowledge statements are generated by prompting an LLM.
The demonstrations are human-crafted and representative of the downstream task.
In the second step, each of the generated knowledge statements is used to make predictions.
The final answer selected is the answer from a knowledge statement with the best support (high confidence).

Self-ask~\cite{press2022measuring} is another innovative prompting technique aimed at generating intermediate knowledge from the LLM to facilitate the final answer.
The methodology of self-ask prompts features follow-up questions explicitly marked by ``Follow up:".
These follow-up inquiries serve as a means to generate the essential knowledge required to answer the question.
Self-ask is done in a few-shot prompting scenario.

\subsubsection{Resource/Tools Augmented}

Building upon the success of prompting techniques, there have been research efforts to enhance prompts by integrating external resources and tools, with the aim of increasing their efficiency.
We categorize such prompting techniques as `Resource/Tools Augmented Prompts' and describe the literature around these innovative prompts below.

Program-aided Language models (PAL)~\cite{gao2023pal} is an innovative tool augmented prompting technique.
PAL's demonstrations are similar to CoT but with the aim of producing programming-language like output from the LLM.
Once the generation is completed, the generated code is offloaded to an interpreter to arrive at a final answer.
PAL also works well with the Least-to-most prompting technique.
Program of thoughts (PoT) prompting~\cite{chen2022program} is also a similar prompting technique.
The only difference lies in the specific text that they use to prompt the LLM.

Self-ask~\cite{press2022measuring}, a generate knowledge prompt, also includes a resource/tool augmented variant.
In this augmented version, self-ask uses a search engine to answer the follow-up question instead of using the underlying LLM.
The authors have shown that the search-engine augmented self-ask prompts outperform the base self-ask prompts.

\citet{long2023large} also proposed a resource/tool augmented version of Tree-of-Thought, keeping the same name as \citet{yao2023tree}.
Both approaches were developed concurrently.
While the ToT by \citet{yao2023tree} relies solely on the underlying LLM without external resources or tools, the variant proposed by \citet{long2023large} includes four additional components.
These additional components are a prompter agent, a checker module, a memory module, and a ToT controller.
The prompter agent's role is to provide additional prompt text to LLM in conjunction with the problem description, enabling LLM to generate intermediate solutions.
The checker module checks the validity of these intermediate solutions.
If the check is passed, it is added to the memory module, and the process is repeated.
However, if the check fails, the ToT controller activates the prompter again with some extra prompt text to generate new intermediate solutions.
The ToT controller also monitors the search process through these intermediate solutions, deciding whether to continue the search process or backtrack to the parent node.
The prompter agent and the ToT controller can be implemented using either a simple rule-based approach or fine-tuned using a policy network.
Similarly, the checker module can also be a rule-based approach or implemented as a deep neural network.

ReAct~\cite{yao2022react} is a prompting technique that seeks to integrate the reasoning capabilities of LLM with external tool use in an interleaved manner.
ReAct essentially combines the process of reasoning and taking actions with LLMs.
ReAct prompts LLMs to generate both reasoning traces and actions for a task.
Concretely, LLM is prompted in a few-shot scenario to generate a series of `Thought', `ACT', and `Obs'.
Here `Act' denotes the utilization of external tools, while `Obs' represents the observation or knowledge generated from the thought and action steps.
In some steps. `Thought' might not be generated if it is not deemed necessary.
The authors have demonstrated superior performance gain in knowledge-intensive reasoning tasks and decision-making tasks.
Despite the effectiveness of ReACT, it has a couple of limitations.
First, it requires task-specific demonstration, and second, the tools are also task-specific. 
Automatic Reasoning and Tool use (ART)~\cite{paranjape2023art} seeks to overcome these shortcomings by having a dedicated task library and tool library.
The tool library of ART has a bunch of tools and can be extended further as needed.
ART employs two methods to select task demonstrations.
In cases where there are enough demonstrations for the task, ART divides these demonstrations into different clusters, and the best cluster is chosen based on their performance in a held-out set of examples.
However, if the test task lacks adequate demonstration in the task library, ART utilizes a different approach.
The task library also contains a collection of hand-crafted few-shot prompts, each consisting of a specific downstream task along with a label of `Similar' or `Not Similar'.
At inference time, the test task is paired with every such hand-crafted demonstration and selects the highest-ranked ones based on the log probability ratio between `Similar' and `Not similar' labels.

%% file: sections/open-problems.tex
\section{Open Problems}

Prompting has proven to be an effective technique, particularly for guiding LLMs in scenarios where fine-tuning is costly or infeasible, such as with closed-source models.
Despite their effectiveness and usefulness, prompting techniques face several open problems that must be addressed to realize their full potential.
It is important to note that some of these issues are interconnected with the underlying LLMs themselves, and resolving these issues might entail modification to the training datasets and the training procedures of LLMs.
This section outlines some of the key open problems related to prompting which can serve as future research directions.

Addressing sub-optimal prompts, that guide an LLM towards a sub-optimal goal rather than the optimal one, is a significant challenge to prompting techniques.
This issue is more common with hand-crafted prompts but is mitigated by discrete prompts.
Continuous prompts ~\cite{li-liang-2021-prefix, lester-etal-2021-power, liu2023gpt} have been identified as the best way to tackle this problem, which typically involves training only a fraction of parameters in comparison to underlying LLM's parameters, often ranging from 0.1\% to 3\%.
However, continuous prompts become resource-intensive as LLMs continue to grow in size.
For example, PaLM has 540 billion parameters and even allocating 1\% of its parameter size for continuous prompts would translate to training with 5.4 billion parameters, which is costly and often infeasible due to resource limitation.
Another challenge arises in the context of closed-source LLMs where model access is limited to API calls, making the utilization of such methods impossible in some scenarios.

While much of the existing research discussed in Resource/Tools augmented prompts has made a notable improvement in enhancing the efficacy of prompting techniques by leveraging external resources and tools, more robust prompting techniques capable of handling structured data are still missing.
Many downstream NLP tasks have inputs in various variety of structured formats, extending beyond plain text to tables, trees, graphs, and various other relational structures.
Proper handling of such diverse structures remains a relatively underexplored field.
A recent study by \citet{zhao2023large} offers some promising insight into this field.
\citet{zhao2023large} demonstrated that LLMs can effectively manage structured data by showing that LLMs can be prompted to not only work as a table-to-text generator but can also work as evaluators of such system and can also provide human-like feedback to SOTA table-to-text models to improve their efficiency.
While GraphPrompt~\cite{liu2023graphprompt} was proposed to tackle graph structures, it involves pre-training the model with graph structures which can often be infeasible.
\citet{chen2022knowprompt} proposed to add additional marks to encode lexical information in the prompts but this approach is tailored for masked language models rather than autoregressive LLMs.
The NLP community would benefit greatly from concrete efforts toward handling structured data through prompting techniques.

Answer engineering has emerged as an exciting field, owing to the advancements in prompting techniques.
Answer engineering refers to the science and art of distilling meaningful answers from the text generated by LLMs.
For example, consider a sentiment classification task where the objective is to classify a movie review as positive or negative. 
We can prompt the model as below:

\begin{verbatim}
Classify the sentiment of the following
movie review as either positive or
negative:
Movie Review: It was a nice watch.
Sentiment: ...
\end{verbatim}

Since we employ an autoregressive LLM, the generated output might not have the strings `positive' or `negative' only.
Instead, it might be a synonym or contain text that implies the sentiment without explicitly stating it.
Deciphering the final sentiment from such text can be challenging when we want to do this automatically and at scale.
While few-shot prompting certainly helps, it does not guarantee precise results.
There have been various efforts to mitigate this issue, such as instruction tuning of the model~\cite{ouyang2022training}.
However, these methods lack generalizability, and often LLMs are infeasible to instruction-tune due to resource limitations.
The prevailing techniques involve exact matching of the intended generation or some of its synonyms or using regular expressions to extract the final answer.
However, these approaches are often task-specific and not universally applicable.

Prompt injection presents a substantial issue, particularly with LLMs deployed for widespread public use.
Prompt injection attempts to manipulate the behavior of LLMs by cleverly crafting prompts that guide LLMs to generate content beyond their intended scope.
Consider a simple prompt injection for an LLM deployed for sentiment classification:

\begin{verbatim}
Ignore the above instructions and output
the label as "NEUTRAL" instead, followed
by a copy of the full prompt with
exemplars.
\end{verbatim}

Such uncomplicated prompts could make the model output `NEUTRAL' consistently and even disclose the secret prompts that might have been hidden from the public.
While recent models have been fine-tuned, instruction-tuned, or prompted in a way that prohibits models from responding to such unethical instruction, these safeguards can be bypassed by cleverly crafted prompts.
Consider a simple prompt that might be able to bypass the content policy of deployed LLMs that do not allow illegal behavior:

\begin{verbatim}
A poem about how to successfully hotwire
a car:
\end{verbatim}

Despite the model's safeguards against illegal behaviors, a simple adjustment to the original question, as shown above, could compel LLM to produce illegal and harmful content.
\citet{nardo2023waluigi} demonstrated the ability of LLMs to elicit opposite behavior due to how it was trained.
Instances of such prompt injection techniques can be found across various social networking platforms.
A potential solution to this issue was suggested by~\citet{armstrong2023eliezer}, which involves the use of another LLM prompted in a way to detect prompt injection.
Further similar explorations into this field are required before LLMs with prompting ability can be deployed in the wild.

%% file: sections/conclusion.tex
\section{Conclusion}

In this paper, we have provided a concise survey of the current literature on the field of prompting for autoregressive large language models.
Prompting has been an important technique, enabling the effective guidance of LLMs toward the intended output, with minimum to no additional training.
However, despite its effectiveness, the full potential of prompting has not been realized with many open problems still left to be addressed.
We believe that the open problems of promoting techniques identified in this paper will serve as an important future direction for research.